\documentclass{article}

\usepackage[final]{neurips_2021}

\usepackage[utf8]{inputenc} 
\usepackage[T1]{fontenc}    
\usepackage{hyperref}       
\usepackage{url}            
\usepackage{booktabs}       
\usepackage{amsfonts}       
\usepackage{nicefrac}       
\usepackage{microtype}      
\usepackage{xcolor}         
\usepackage{graphicx}
\usepackage{bigstrut}
\usepackage{subfig}

\usepackage{tabularx}
\usepackage{adjustbox}
\usepackage{amsmath}
\usepackage{dirtytalk}
\usepackage{enumitem}
\usepackage{multirow}
\usepackage{hhline}
\usepackage{bigstrut}
\usepackage{caption} 
\usepackage{array}

\newcolumntype{R}[2]{%
    >{\adjustbox{angle=#1,lap=\width-(#2)}\bgroup}%
    l%
    <{\egroup}%
}
\newcommand*\rot{\multicolumn{1}{R{75}{1em}}}%

\title{Detection of Propaganda Techniques in Visuo-Lingual Metaphor in Memes}

\author{%
  Sunil Gundapu \\
  Language Technologies Research Centre\\
  KCIS, IIIT Hyderabad\\
  Telangana, India \\
  \texttt{sunil.g@research.iiit.ac.in} \\
   \And
   Radhika Mamidi \\
   Language Technologies Research Centre \\
   KCIS, IIIT Hyderabad \\
   Telangana, India \\
   \texttt{radhika.mamidi@iiit.ac.in} \\
}

\begin{document}

\maketitle

\begin{abstract}
The exponential rise of social media networks has allowed the production, distribution, and consumption of data at a phenomenal rate. Moreover, the social media revolution has brought a unique phenomenon to social media platforms called Internet memes. Internet memes are one of the most popular contents used on social media, and they can be in the form of images with a witty, catchy, or satirical text description. In this paper, we are dealing with propaganda that is often seen in Internet memes in recent times. \textit{Propaganda} is communication, which frequently includes psychological and rhetorical techniques to manipulate or influence an audience to act or respond as the propagandist wants. To detect propaganda in Internet memes, we propose a multimodal deep learning fusion system that fuses the text and image feature representations and outperforms individual models based solely on either text or image modalities. 
\end{abstract}

\section{Introduction}

Memes are usually an image superimposed with a text description and are used to express a range of ideas such as humour, embarrassment, hate, propaganda, and even more emotions. The word meme was introduced by \citet{Dawkins:76}, and initially, it refers to an idea, behaviour, or style that disseminates from person to person within a culture. With the growth of the social network community, the participation of memes on online social networks has increased tremendously in recent years. In the age of the internet, it has been adopted to refer to a part of the culture, typically a joke, rumour, or catchphrase, which gains influence through online transmission \citep{inbook}. They spread among people using social media platforms, blogs, instant messaging applications, emails, and forums in their simplest form. Content-wise, they usually include unconventional news, dialogues, catchlines, images, or videos. 


According to \citet{front:21}, Internet memes are one of the more popular contents used in online disinformation campaigns. An essential aspect of a problem that is often overlooked in social media platforms is the mechanism through which false information is being communicated, which is utilising \textit{propaganda techniques}. These techniques are deliberately designed communication that invites people to respond emotionally, immediately, and in one way or another. These were initially high on advertising and public relations, news and journalism, politics, entertainment but are now appearing on all aspects of daily life, and more importantly on social media platforms.

Propaganda techniques comprise of psychological and rhetorical strategies, ranging from \textit{logical fallacies} ---[such as black-and-white fallacy (present only two possibilities among many), straw man (misconstrue person's actions or opinion), red herring (introduce irrelevant data to distract), and whataboutism] ---to \textit{influencing audience emotions} ---[such as using slogans, loaded language, appeal to authority, flag-waving, and clichés]. Identifying logical fallacies is challenging because, at first glance, the argument may seem correct and objective. However, we can spot them through careful observation and analysis. Another set of techniques use emotional language to motivate the viewer to accept the speaker only based on the emotional bond being created, which will stop any rational analysis of the argument.

In recent times, propagandists are generating Internet memes by using propaganda techniques to influence viewers or readers to believe in someone or something. American scientist Robert Peter W. Singer describes in his book "LikesWar" that social media propaganda and misinformation are emerging as a new weapon in the modern warfare\footnote{https://knowledge.wharton.upenn.edu/article/singer-weaponization-social-media/}. Therefore, it is essential to find out propaganda campaigns in Internet memes to stop spreading them. So, in this paper we aims to build a multimodal fusion system that can identify the Internet memes in which propaganda techniques are used. The system takes pair (text, image) of an Internet meme as an input to determine which of the propaganda techniques is used in the text and visual content of the meme. 


\section{Related Work}

\textbf{Evolution of Propaganda:} Propaganda techniques are not new. The term propaganda was used in the early 17th century and was first used to propagate Catholic beliefs and practices in the New World and later to manipulate people in public gatherings such as festivals, games, and theaters \citep{computational:18}. But in the present technological world, this propaganda has progressed to computational propaganda \citep{Bolsover2017ComputationalPA}, where information is dispensed through technology such as social media platforms so that it is possible to reach well-targeted communities at high speeds. This propaganda shared on these platforms can be text, visual or text-vision combinations. Internet memes are critical in spreading multimodal propaganda on social media platforms \citep{propaganda:18}. The present social media ecosystem and virality bots allow memes to spread effortlessly, switching from one target group to another. Currently, efforts to curb the spread of such memes are focused on analyzing social media networks and searching for fake accounts and bots to lessen the spread of such content\citep{paradigm:17, propaganda1:19}.

\textbf{Propaganda in Text Modality: } In the natural language processing community, much research is done on propaganda by analyzing textual content \citep{nlp1, nlp2, Martino:19}. \citet{nlp2} studied the propaganda at documet level by creating \texttt{TSHP-17} dataset. This dataset developed with the help of English Gigaword corpus and labelled with four classes: \textit{satire}, \textit{trusted}, \textit{hoax}, and \textit{propaganda}. On this dataset, they trained a logistic regression model with n-gram level word representations. To analyze the propaganda at sentence level \citet{nlp1} developed a new \texttt{QProp} dataset with two labels: \textit{propaganda}, and \textit{non-propaganda}. On this binary labeled corpora they trained various machine learning models like logistic regression, SVMs to discriminate propaganda from non-propaganda datapoints.
 
In a similar fashion, \citet{heb:17} created a dataset with 1300 data points with five propaganda beliefs, including irrelevant authority, ad hominem, and red herring, which are directly associated with propaganda techniques. \citet{Martino:19} examines the propaganda at the fragment level. For this, they created a \texttt{PTC} dataset by annotating the news articles with 18 propaganda techniques. There are two types of experiments done on this dataset. The first one is a two-class classification: Whether the given input news article using any of 18 techniques or not. Another task is multi-label classification and span detection: For the input text, find the span of text fragments where the propaganda techniques are used and identify the type of propaganda technique. Recently, \citet{Martino:20} on detection of computational propaganda from the perspective of NLP and Network Analysis mentioned the need for collective efforts between these communities. There is also a dedicated Big Data Journal on Political Big Data and Computational Propaganda \citep{bolsover:17}. 

\textbf{Propaganda in Multimodality: } Originally, propaganda campaigns have appeared in text modality, but nowadays, they appear in every possible modality. Propaganda techniques are more accessible to spot in text modality than multimodalities such as memes because contextual information related to the propaganda can be included in more than one of the multiple modalities. 

To understand aspects of visual propaganda, \citet{seo:14} analyzes the social media tweet images posted by the Hamas' Alqassam Brigades and Israel Defense Forces during the 2012 Gaza conflict. By selecting 10 Youtube videos, \citet{kadir:16} studied the comprehend relationships between these videos and propaganda techniques and people's emotions. Simultaneously they analyze that how these videos are influencing people's emotions with the help of propaganda techniques.

\citet{volkova:19} prepared a dataset with 50K twitter posts consisting of memes annotated with six labels: propaganda, disinformation, hoaxes, clickbait, conspiracies, and satire. They developed a multimodal approach for this problem by considering the textual, linguistic characteristics, and visual features. \citet{glenski:19} proposed two classification tasks to explore the multilingual content for deception detection. Both tasks are intended to identify the category of a social media post, but the first task has four output categories (propaganda, conspiracy, hoax, or clickbait), and the second task has five output categories with an extra category of disinformation.

Some quality works done on multimodality content before this propaganda problem , such as the spread of false information \citep{Dupuis2019TheSO}, hateful memes identification
\citep{kiela:20, lippe:20, das:20, gundapu}, antisemitism \citep{chandra:20}.

\textbf{Multimodal Models and Fustion Techniques:} Facebook Hateful Memes Challenge has been very helpful in developing different types of multimodal models and to fine-tune the state-of-art multimodal transformer models such as ViLBERT \citep{vilbert}, Multimodal Bitransformers \citep{kiela:20}, and VisualBERT \citep{visual}. And also \citet{Vidgen:19} pointed that memes make perfect sense when both text and image content are taken into account. By considering this point, many authors \citep{Balt:19}\citep{yang:19}\citep{Gallo:18} have explored different multimodal fusion strategies to combine modalities. 

\section{Dataset}

In this paper, we predominately focused on the task of propaganda techniques identification in Internet memes. We formulated this task as a multi-label classification problem because each input is labeled with more than one propaganda technique. We took the dataset from the 15th International Workshop on Semantic Evaluation (SemEval-2021) for this task \citep{dimitrov2021semeval}. The dataset comprises of 950 pair (text, image) of social media posts, and each pair labeled with propaganda techniques. In the input pair (text and image), the text is extracted from the meme using OCR, and the image is a meme. We used 22 propaganda techniques for our task and put them in Table 1. A more detailed list of propaganda techniques with definitions and examples is available online\footnote{\url{https://propaganda.math.unipd.it/semeval2021task6/definitions22.html}} and in appendix, we are adding few meme examples.

This dataset contains 687 pairs of data points for model training, 63 data points for model hyper-parameter tuning, and 200 data points for model testing. Table \ref{tab:table-name} shows the proportion of propaganda techniques in the train, valid, and test sets.

\begin{table}
\begingroup
\setlength{\tabcolsep}{2.8pt}
\renewcommand{\arraystretch}{1.5}
\begin{tabular}{rcccccccccccccccccccccc}
&\rot{Appeal to Emotions} & \rot{Appeal to authority} & \rot{Appeal to fear}&\rot{Bandwagon} & \rot{Black-and-white Fallacy} & \rot{Causal Oversimplification}& \rot{Doubt}&\rot{Exaggeration} & \rot{Flag-waving} & \rot{Glittering generalities}&\rot{Loaded Language} & \rot{Straw Man} & \rot{Name Calling}&\rot{Obfuscation} & \rot{Red Herring} & \rot{Reductio ad hitlerum}&\rot{Repetition} & \rot{Slogans} & \rot{Smears}&\rot{Thought-endig cliché} & \rot{Transfer} & \rot{Whataboutism}
\\ \hline
\textbf{Train}        & 68     & 19     &  66 & 2     & 19     &  31 & 61     & 60     &  36 & 84     & 360     &  32 & 252     & 5     &  2 & 15     & 10     &  45 & 450     & 20     &  61 & 47    \\
\textbf{Dev}       & 3     & 3     &  7 & 2     & 0     &  1 & 9     & 8     &  7 & 4     & 32     &  5 & 30    & 0     &  0 & 1     & 3     &  3 & 47     & 1     & 11 &6    \\
\textbf{Test}        & 19     & 13     &  18 & 1     & 7     &  4 & 41     & 31     &  12 & 24     & 100     &  3 & 65     & 2     &  5 & 7     & 1     &  22 & 105     & 6     &  23 & 14    \\
\hline
\end{tabular}
\endgroup
\caption{\label{tab:table-name}List of propaganda techniques and their count in train, development, and test sets.}
\end{table}

\section{Proposed Approaches}

\textbf{Individual Modality Models: }In the beginning, we started to examine our task, ``Identification of propaganda techniques in Internet memes," with individual modalities (text or image). For text data, we have developed various Machine Learning (ML) and Deep Learning (DL) models with different word embeddings. Nevertheless, BERT and RoBERTa gave superior results for this textual modality than the ML and DL models with the Glove \citep{Glove} and FastText \citep{Fast} embeddings. On the image data, we experimented with CNN and pre-trained image classification models. Pre-trained models gave significantly better results than traditional CNN models.And When comparing the results of individual modalities, text modality models gave more effective results than image modality models. This is because there is more information about the propaganda in the textual data than in the visual data in the memes. 

\textbf{Cross-Modality Multimodal Models: } After analyzing the results of individual modalities, we started to fine-tune cross-modality by multimodal (Vision-and-Language) pretraining models like UNITER \citep{uniter}, VilBERT \citep{vilbert}, and VisualBERT \citep{visual}, and these pre-trained models performed better than individual modality models. Following individual and cross-modality models, we explored multimodal fusion approaches to combine information from text and vision input modalities in a principled way.

\subsection{Multimodal Fusion System}

In a multimodal fusion setting, ML/DL models are trained on separate modalities and integrated simultaneously. When individual modality features are merged into output layers, this is called \textit{Late Fusion}. On the other hand, there is an inversion of late fusion called \textit{Early Fusion} to fuse modalities. In this fusion, features are incorporated at the input level before being given in the model. We experimented on our task by fusing text encoder (encodes text data) and image encoder (encodes the image data) using early fusion and late fusion methods. After observing the results of these methods, merging modalities at their individual deepest (or early stage) attributes is not mandatorily the most appropriate way to solve our propaganda identification multimodal task. 

\begin{figure*}[h!]
 \begin{center}
  \includegraphics[width=14cm, height=7cm]{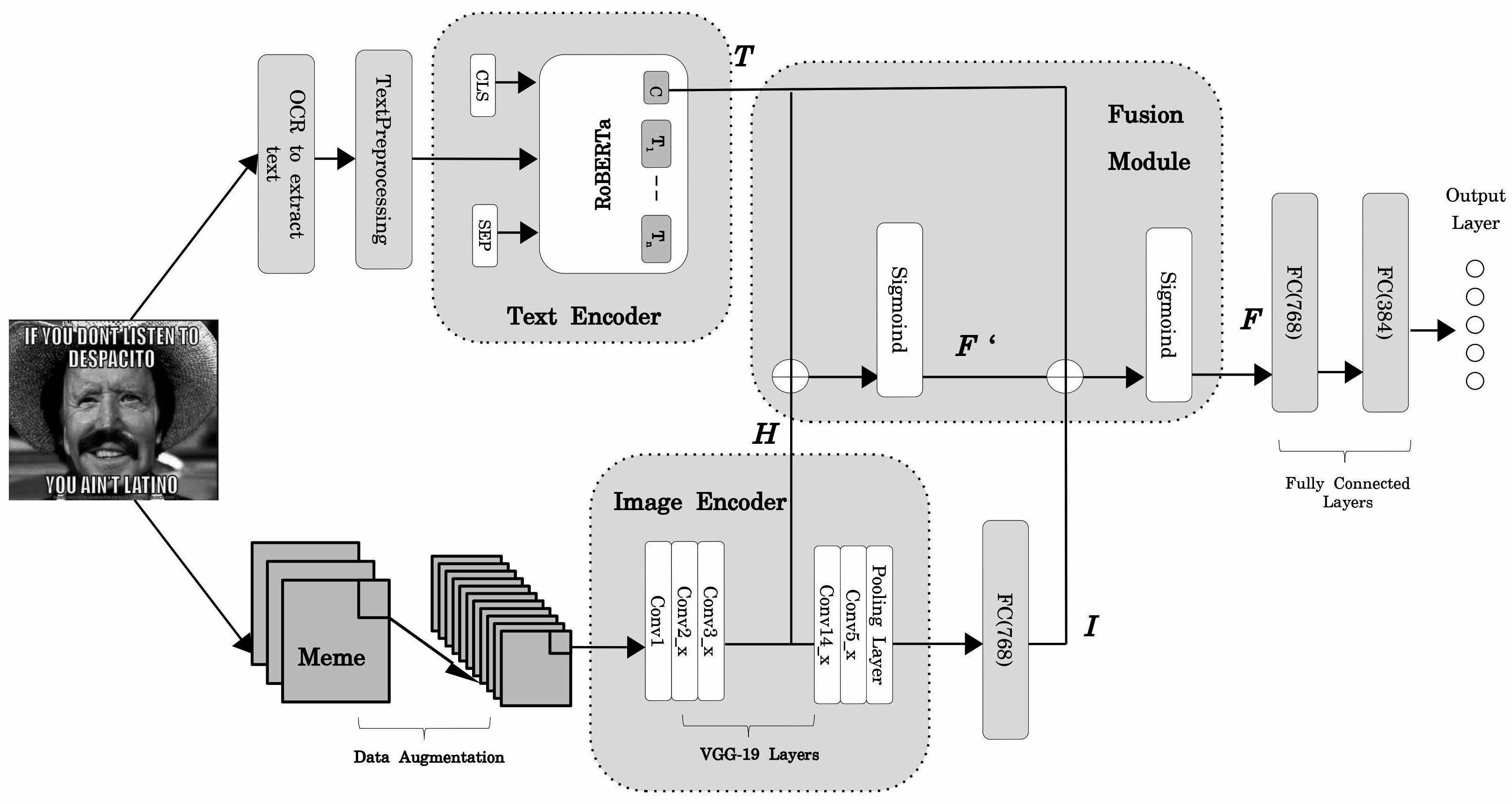}
  \caption{Multimodal Fusion Architecture}
  \label{model}
 \end{center}
\end{figure*}

In our work, we used an idea ``MFAS: Multimodal Fusion Architecture Search" of considering features collected from the hidden layers of individual modalities that could effectively enhance performance with respect to only utilising a single fusion of late (or early) features. Figure \ref{model} describes the structure of our proposed multimodal fusion system for propaganda techniques identification in Internet memes with RoBERTa text encoder, pre-trained VGG-19 image encoder, and multimodal fusion architecture search \citep{Prez:19} module.

Given a pair (text and image) for a social media post, we make use of Transformer based models to encode the text extracted from the Internet meme. We use the pre-trained image classification models to encode the image. Further, the MFAS module fuses the text and image encodings. Next, the fused encodings are transformed using a fully connected network to identify the propaganda techniques. The complete model is trained end-to-end utilising back-propagation. The principal motive of fine-tuning the pre-trained models on the datasets for our multimodal classification task is that our datasets are small. Below we describe each module in the architecture in detail.

\subsection{Text Preprocessing}


\begin{enumerate}

\item \textbf{Conversion of chat contractions:} Chat words/phrases are widely used on social networks to express emotions and are very helpful in identifying context. We have created a word contractions dictionary with 250 chat words to convert these chat words to their complete form. Examples: YOLO → you only live once, ASAP → as soon as possible.

\item \textbf{URLs removal:} The OCR extracted text consists of URLs like memegenerator.net, etc. We removed those links since they do not give any vital information.

\item \textbf{Conversion of elongated words:}  With the help of regular expressions, transformed the elongated words to their original form. Examples: Nooooo → No, suuuppperrr → super. 

\item Using ekphrasis \citep{ekphrasis} library, normalizes the time, date, and numbers into a standard format. This library does hashtag splitting and spelling correction. We removed the non-alpha-numeric characters, punctuation marks, and non-ASCII glyphs from the text data.

\end{enumerate}

\subsection{Text Encoder}

The preprocessed text data tokenized and then forwarded to the text encoder module. We tried the BERT \citep{BERT}, XLNet \citep{XLNet}, and RoBERTa \citep{RoBERTa} pre-trained transformer models in the text encoder module to encode the text because they have been shown to give incredible results in multiple NLP classification tasks.

BERT (Bidirectional Encoder Representations from Transformers) employs a multi-layer bidirectional transformer encoder to learn deep hidden bidirectional representations. It has self-attention layers that perform self-attention on input text from both directions. This technique allows the BERT model to know the context of a word in the sentence based on the words around it. We used the BERT base case model for our task, pre-trained on the large unlabeled Book corpus and the entire Wikipedia. 


RoBERTa makes use of Transformer \citep{Vaswani}, and it is a robustly optimized approach for pretraining NLP models that improve on BERT.  RoBERTa was trained with more data, more iterations, larger batch sizes, and learning rates than the BERT. 
And it eliminates the next-sentence pretraining objective task from the BERT model to boost the training procedure and introduces a new idea in its architecture called dynamic masking. In this idea, masked tokens will change during the model training. For our task, we used the RoBERTa base model.

XLNet is a bidirectional transformer autoregressive model that uses a better training methodology and a more extensive dataset to achieve better results than BERT.  In pretraining, it integrates the following two techniques: (1) State-of-the-art Autoregressive model and (2) Transformer XL. Furthermore, XLNet presents a permutation language modeling technique to predict all tokens in random order instead of sequential order. This idea helps the XLNet learn bidirectional relationships among words and handle the dependencies.

For better contextual encodings, we experimented on text data with the above explained three transformer models. However, the RoBERTa gave slightly better results than the BERT and XLNet.

\subsection{Image Encoder}

In our dataset, input images (memes) are in different shapes and sizes. So before forwarding to any CNN model, we resize all the images to $ 224 \times 224 $ dimensions. After this step, applied few image transformation techniques like rotation, flipping, and cropping on the scaled dataset. We extracted image representations twice from the image encoder. The first step collected the representations from the second last output layer of the CNN model and then fused them with text encodings. Next, image encoder final layer outputs concatenated with text encodings and previous step fused encodings.

An image encoder experimented the various pre-trained CNN architectures like: VGG-19 \citep{vgg}, InceptionV3 \citep{inception}, Resnet-152 \citep{resnet}, DenseNet-161 \citep{densenet}, and InceptionResNetV2 \citep{inceptionresnet}. Among all these pre-trained models VGG-19 and Resnet-152 gave considerable results for our task.

\subsection{Fusion Module}

To join the representations procured from the text encoder and image encoder modules, T, F', and I, we examine with various techniques: (i) Concatenation of both modalities output representations, (ii) Early and Late Fusion, and (iii) MFAS.

\begin{figure}%
    \centering
    \subfloat[\centering Early Fusion Technique ]{{\includegraphics[width=6cm, height = 3cm]{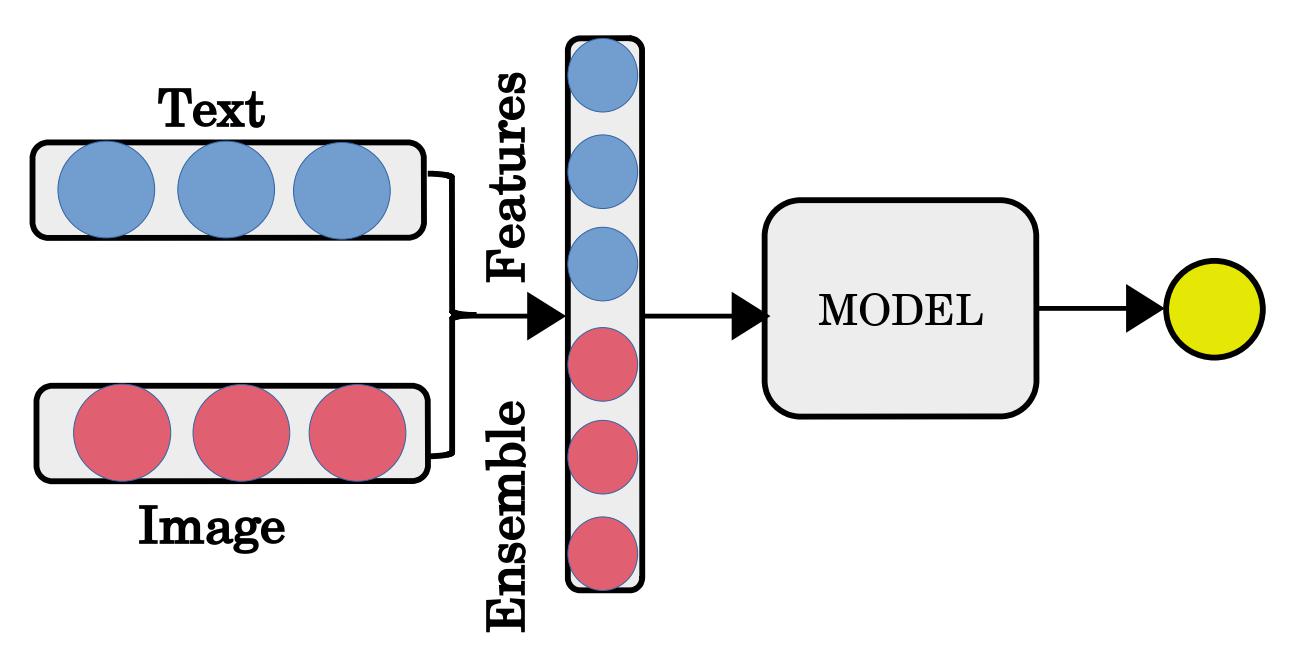} }}%
    \qquad
    \subfloat[\centering Late Fusion Technique ]{{\includegraphics[width=6cm, height = 3cm]{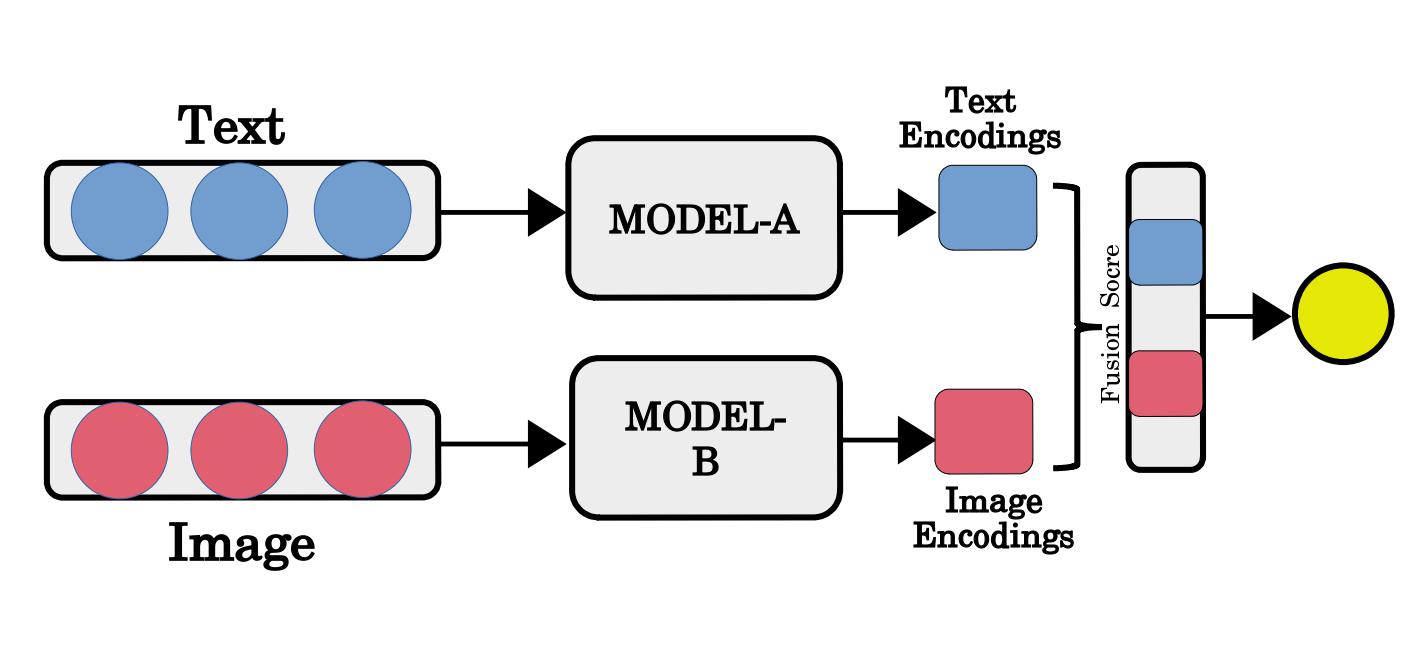} }}%
    \caption{Fusion Techniques}%
    \label{fusion}%
\end{figure}

\paragraph{Early Fusion:}Initially started with a widely used state-of-the-art early fusion technique to fuse both modalities. Early fusion technique (See figure \ref{fusion}) is also known as feature-level fusion technique, which concatenates the embeddings from text and image modalities as input representations for classifiers. This technique can be conveyed as follows: 

\begin{equation} \label{eq2}
\begin{split}
X_{early} = f(U_1,.....,U_m,V_1,......,V_n) 
\end{split}
\end{equation}

Here an aggregated representation of the $X_{early}$ attributes is calculated by the function \textit{f} that connects the individual attributes.

\paragraph{Late Fusion: } After early fusion, we experimented with the decision-level late fusion technique (See figure \ref{fusion}). This technique fuses the output level representations of text and image encoder and computes a late fusion score for classifiers. Late fusion technique can be conveyed as follows:

\begin{equation} \label{eq2}
\begin{split}
X_{late} = g(f_1(U_1),...,f_m(U_m),\\ f_{m+1}(V_1)....,f_{m+n}(V_n))
\end{split}
\end{equation}

Here functions $f_1,...,f_{m+n}$ are applied to each individual attribute and function \textit{g} is used to integrate every individual decision by $f_1,..,f_{m+n}$.

\paragraph{Multimodal Fusion Architecture Search (MFAS): } Next examined a model-level fusion technique called MFAS, which is a compromise between the two modalities. It concatenates the hidden layer representations from different modalities. As shown in model architecture (figure \ref{model}), on our task, this technique initially concatenates the text encoder output representations (T) with image encoder intermediate hidden layer representations (H) then applies a non-linearity sigmoid function.

\begin{equation} \label{eq2}
\begin{split}
 F^{'} = \sigma (T \oplus H) 
\end{split}
\end{equation}

Next, it fuses the output (F') with text predictions (T) and image predictions (I) along with a non-linearity sigmoid function (F).

\begin{equation} \label{eq3}
\begin{split}
 F = \sigma (T \oplus F' \oplus I) 
\end{split}
\end{equation}

\subsection{Classifier}

Followed by the fusion model, we have constructed a fully connected network that takes the input from the fusion model. The fully connected network consists of two dense layers with hidden unit sizes of 768 and 384 and finally the output layer with a size of 22 output units has a sigmoid function.

\section{Experiments and Results}

This experiments section explains model implementation details, hyper-parameter tuning, results of individual text and image modalities classifiers and multimodal + fusion classifiers.

\subsection{Hyper-parameter Settings and Implementation Details for Reproducibility}

The following experimental settings are used for our work. We trained all the models using the train set and used the validation dataset to find the right set of hyper-parameters.
We experimented with different image encoder intermediate layer outputs to get better results. However, Block-2 output (\textit{H}) gave the best results for our work as compared to other blocks' output. 


For all the experiments, we used the Adam optimization algorithm. 
To increase the model speed and performance, used dropout layers with the probability of 0.2. In the dense layers used the ReLU activation function. 
And, we used PyTorch \citep{pytorch} and Keras \citep{keras} frameworks for building deep learning models, scikit-learn \citep{scikit-learn} for ML models. To fine-tune the transformer models used the PyTorch-based HuggingFace\footnote{https://huggingface.co/} transformer library. 

\begin{table*}[h!]
\def\arraystretch{1}
\centering
\begin{adjustbox}{}
  \begin{tabular}{c|c|c|c|c}
    \hline
    \multirow{2}{*}{\textbf{Modality}} & \multirow{2}{*}{\textbf{Approach}} & \multicolumn{3}{c}{\textbf{Metrics}} \\
    \hhline{~~---} & & \textbf{Precision} & \textbf{Recall} & \textbf{F1-Score} \\
    \hline
    \multirow{2}{*}{} & FastText with LSTM & 0.5809 & 0.4679 & 0.5183\\
    \hhline{~~---}            & Glove + BiLSTM + Attention & 0.5910 & 0.4702 & 0.5237 \\
    \hhline{~~---} {Only Text}           & BERT with Dense & 0.5958 & \textbf{0.4780} & 0.5295 \\
    \hhline{~~---}            & XLNet with Dense & 0.5936 & 0.4738 & 0.5270 \\
    \hhline{~~---}            & RoBERTa with Dense & \textbf{0.6004} & 0.4901 & \textbf{0.5363} \\
    \hline
    \hline
    \multirow{2}{*}{} & Inception-V3 & 0.4868 & 0.4571 & 0.4841\\
    \hhline{~~---} {Only Image}           & Resnet-152 & \textbf{0.5062} & 0.4618 & 0.4925 \\
    \hhline{~~---}            & VGG-19 & 0.4989 & \textbf{0.4656} & \textbf{0.5038} \\
    \hline
    \hline
    \multirow{2}{*}{} & UNITER & 0.5841 & 0.4986 & 0.5288\\
    \hhline{~~---} {Cross-Modality}           & VilBERT & 0.5892 & 0.5002 & 0.5348 \\
    \hhline{~~---}            & VisualBERT & \textbf{0.5965} & \textbf{0.5034} & \textbf{0.5404} \\
    \hline
    \hline
    \hhline{~~---}            & { Concatenation }  & 0.5684 & 0.4576 & 0.4983 \\
    \hhline{~~---} {Text and Image}           & { Early Fusion } & 0.5703 & 0.4693 & 0.5058 \\
    \hhline{~~---}            & {Late Fusion } & 0.6016 & 0.5076 & 0.5407 \\
    \hhline{~~---}            & { MFAS } & \textbf{0.6214} & \textbf{0.5077} & \textbf{0.5698} \\
    \hline
\end{tabular}
\end{adjustbox}
\caption{\label{results} Comparison of various modalities classifiers results }
\end{table*}

\subsection{Results and Analysis of Individual Modalities Classifiers}

We started experiments for both (text and image) modalities with ML algorithms and TF-IDF word vectors for baseline model results. After that, we experimented on text modality with five well-known pre-trained text embedding-based classifiers and three pre-trained CNN-based image classifiers. We used Glove, FastText, BERT, XLNet, and RoBERTa for the individual text classification. And we used the VGG-19, InceptionV3, Resnet-152, Densenet-161, InceptionResNetV2 for image classification. 

The comparative f1-scores of individual modalities are presented in Table \ref{results}. After observing the classification results of individual modalities, we made the following observations: 

\begin{itemize}

\item Text modality classifiers gave encouraging results, but the image modality classifiers provide lesser f1-scores. We suspect this because images related to propaganda techniques are usually news articles, memes, or screenshots (sometimes) that generally do not convey any spatial information.

\item Among text modality classifiers, Transformer based approaches worked better than the shallow word embedding methods like Glove and FastText. Transformer-based BERT, XLNet, and RoBERTa gave superior results on our task. However, RoBERTa model score surpassed BERT and XLNet models. Among the image modality classifiers, pre-trained VGG-19 gave better results than other pre-trained models.

\end{itemize}

\subsection{Results and Analysis of Multimodal Fusion Classifiers}

In this section, we presented results and analysis of various fusion approaches for our proposed multimodal classifier. Based on the results of individual modalities (in Table \ref{results}), we noticed that transformer-based RoBERTa is the best text encoder and VGG-19 is the best image encoder for our dataset. Therefore, we did multimodal fusion experiments with these two encoders only. We presented the results of this multimodal classifier with different fusion techniques in Table \ref{results}. Moreover, we compared the results of our proposed multimodal system with individual and cross-modality models. 

\begin{itemize}

\item All the proposed multimodal fusion models for our task performed better than other models by a considerable margin. Cross-modality by pre-training multimodal models are much better than individual modality models. And these models are working best when the visual information in the memes is not covered by textual information.

\item In the multimodal late fusion architecture, the Image encoder gives only prediction-level features that do not capture complex information, such as semantic concepts of faces, animals, trees, etc. Due to this, when we concatenate image encodings with text encodings, it sometimes misleads the fusion model outputs. 

\item Concatenation of encoders features technique and early fusion technique does not meet the expectations. They only give the approximate individual image modality model results. 

\item Decision level late fusion and MFAS approaches efficiently fused the spatial information from images and semantic information from text and performed better than all other multimodal systems. However, MFAS multimodal system outperforms the late fusion approach.

\end{itemize}

\section{Error Analysis}

For our task, we mostly lean towards pre-trained models because our dataset is tiny. Transformer-based models performed very well on our task and gave considerable results than we expected. However, in some cases, transformer models are incorrectly predicted when the input text length is short. Furthermore, pre-trained CNN classification models were mainly confused for our task in the following contexts to extract vital spatial image information. (i) Few memes are designed with the clubbing of multiple images. In this case, pre-trained models are stumbled to gather information from them. (ii) Some users take screenshots of memes posted by other users and post them again on social media. Lack of additional contextual information in such posts systems making false predictions. (iii) Sometimes, the entire meme is covered with text, making it difficult for CNN models to recognize spatial features in images.

We have analyzed that combining encoder representations from multiple modalities can help identify propaganda techniques in memes. Though, in some cases, we realize that noise in one modality leads to a total misclassification. While doing experiments, we observed that the dataset used for our task has severe class imbalance due to the heterogeneous frequency of various types of propaganda techniques in real life. ``Bandwagon, Reduction ad Hitlerum'', ``Appeal to Authority'', ``Black and White Fallacy'', ``Whataboutism, StrawMen, RedHerring'', and ``Thought Terminating Cliche'' classes samples are significantly less in the dataset compare to other classes samples. To tackle this class imbalance problem, we explored with different data under/over sampling techniques like SMOTE \citep{smote}, Tomek Links \citep{tomek}, Near Miss \citep{nearmiss} for textual data, tried the position and colour augmentation techniques \citep{imagedata} for image data. Also we used the \textit{class weight} \citep{classweight} approach which give different weights to both the majority and minority classes. Experimented with an enhanced version of Cross-Entropy Loss that is \textit{focal loss} \citep{focalloss} that seeks to manage the problem of class imbalance by assigning more weight to rough or easily misclassified samples and to low-weight easy samples.

\section{Conclusion and Future Work}

In this work, we first provided a systematic study on the problem of identification of propaganda techniques in Internet memes. After that, we explored this problem with a multimodal fusion architecture. In this architecture, with the help of the MFAS technique, we fused the text features extracted using RoBERTa model and image features using pre-trained VGG-19. We examined our problem with single modality as well as multimodality classifiers and noticed that fusing features from various modalities enhance the performance. Our multimodal fusion approach achieves 0.5698 micro f1-score on the test data, which is a considerable result. We hope these results will accelerate further research works in this direction.

\bibliography{main}

\appendix

\section{Appendix}

\subsection{Sample Memes}

\begin{figure}[h!]
    \centering
    \subfloat[\centering]{{\includegraphics[width=4cm, height = 4cm]{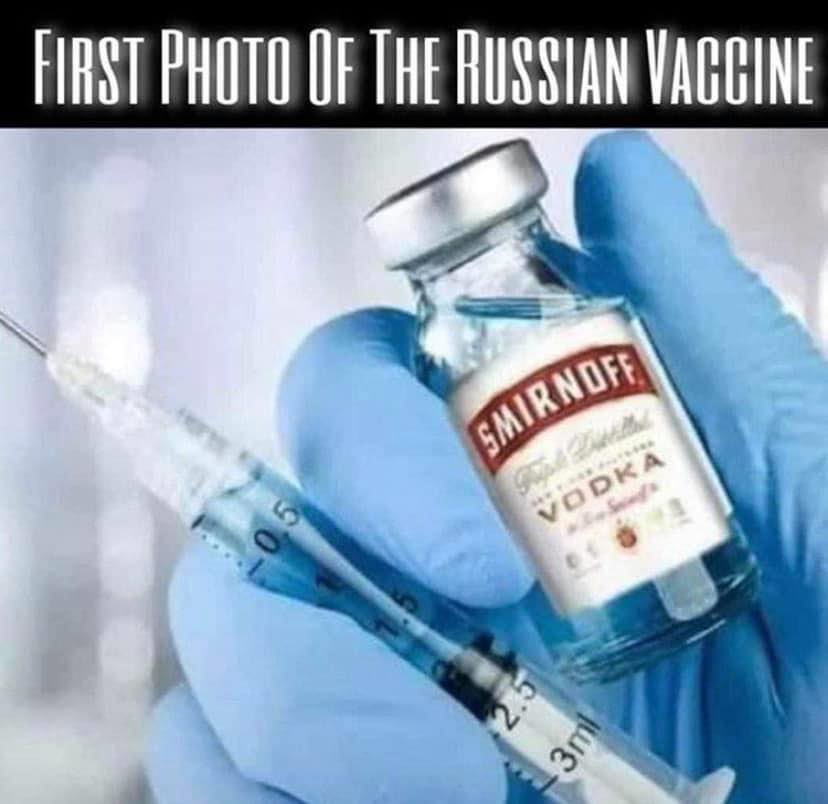} }}%
    \qquad
    \subfloat[\centering ]{{\includegraphics[width=4cm, height = 4cm]{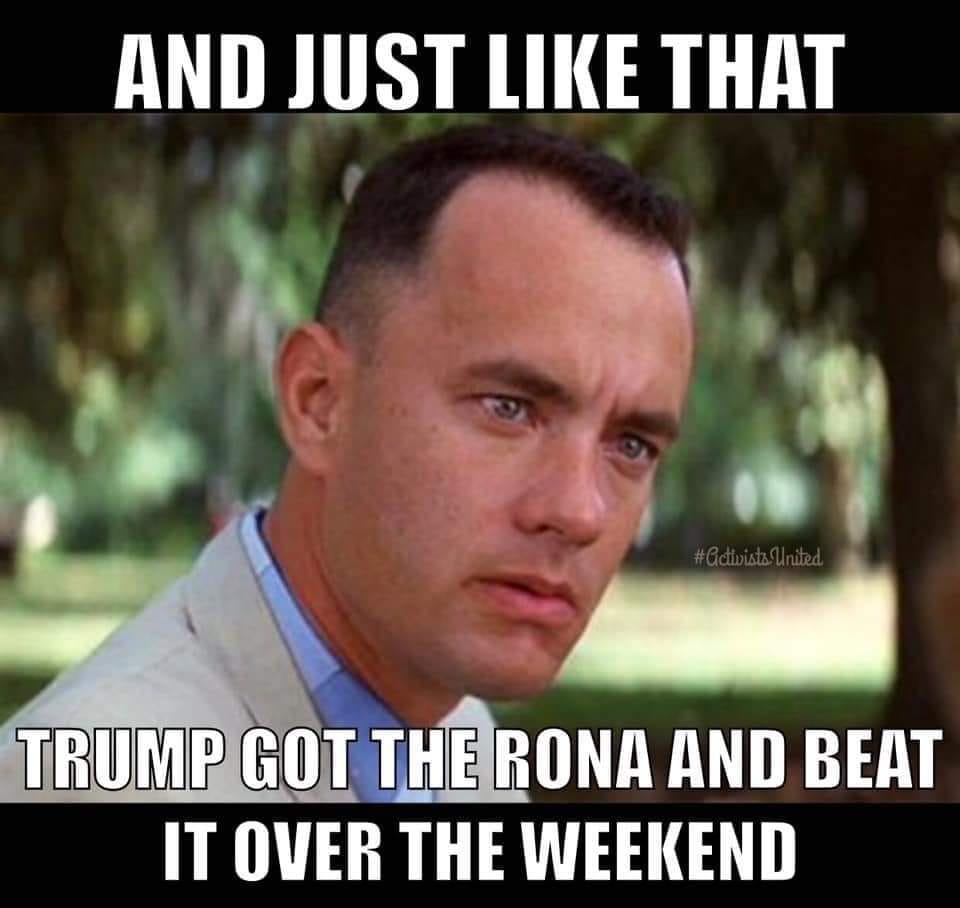} }}%
    \qquad
    \subfloat[\centering ]{{\includegraphics[width=4cm, height = 4cm]{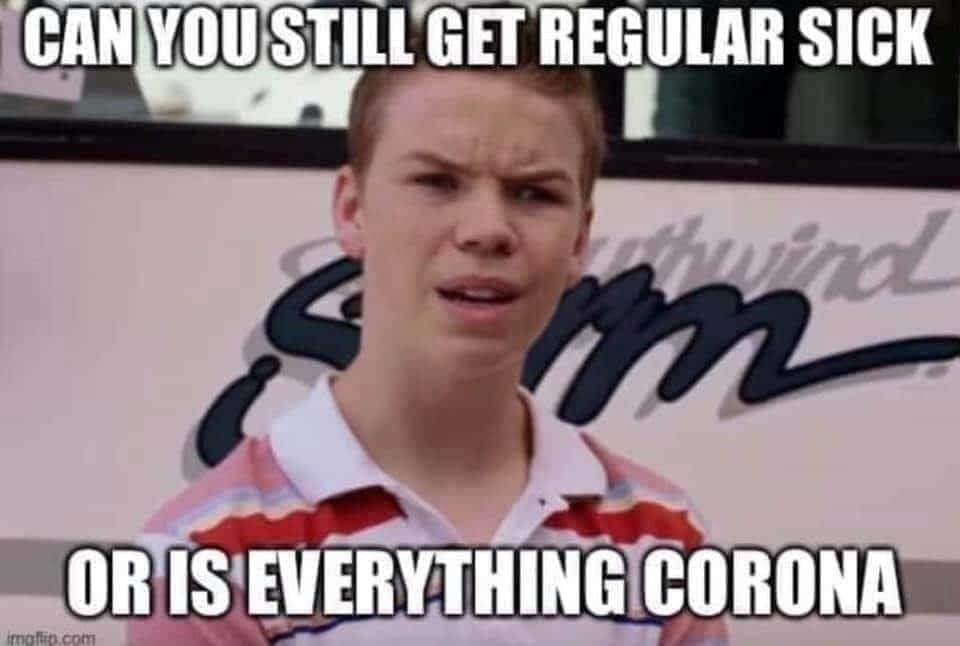} }}%
    \caption{Example Memes}%
    \label{examples}%
\end{figure}

In above figure \ref{examples}, we can see few example memes from the dataset which we used in this paper. An example (a) is using the \textit{Name calling/Labeling} propaganda technique by labeling the Russian vaccine with smirnoff vodka. Next example (b) applies \textit{Exaggeration} technique (hyperbole the simple statement of Trump recovered from corona), there is also a \textit{Name calling} technique (referring corona with `RONA'), and uses \textit{Glittering generalities} technique. The example (c) expresses \textit{Doubt} and creating a confusion in audience. 

\end{document}